\documentclass[runningheads]{llncs}
\usepackage{graphicx}

\usepackage{hyperref}
\hypersetup{
    colorlinks=true,
    linkcolor=blue,
    filecolor=magenta,      
    urlcolor=black,
}

\begin{document}

\title{Deep Visual Attention-Based Transfer Clustering}
\titlerunning{Deep Visual Attention-Based Transfer Clustering}
\author{Akshaykumar Gunari \and
Shashidhar Veerappa Kudari \and
Sukanya Nadagadalli \and
Keerthi Goudnaik \and
Ramesh Ashok Tabib \and
Uma Mudenagudi \and
Adarsh Jamadandi
}

\institute{KLE Technological University, Hubballi, India\\
\email{http://kletech.ac.in/}\\
akshaygunari@gmail.com, shashidharvk100@gmail.com, sukanyanadagadalli@gmail.com, keerthigoudnaik@gmail.com, ramesh\_t@kletech.ac.in, uma@kletech.ac.in, adarsh.cto@tweaklabsinc.com
}

\authorrunning{Akshaykumar et al.}

\maketitle              
\begin{abstract}

In this paper, we propose a methodology to improvise the technique of deep transfer clustering (DTC) when applied to the less variant data distribution. Clustering can be considered as the most important unsupervised learning problem. A simple definition of clustering can be stated as “the process of organizing objects into groups, whose members are similar in some way”. Image clustering is a crucial but challenging task in the domain machine learning and computer vision. We have discussed the clustering of the data collection where the data is less variant. We have discussed the improvement by using attention-based classifiers rather than regular classifiers as the initial feature extractors in the deep transfer clustering. We have enforced the model to learn only the required region of interest in the images to get the differentiable and robust features that do not take into account the background. This paper is the improvement of the existing deep transfer clustering for less variant data distribution.

\end{abstract}

\section{Introduction}

Clustering is central to many data-driven application domains and has been studied extensively in terms of distance functions and grouping algorithms. Relatively little work has focused on learning representations for clustering. One such methodology came up with deep embedded clustering (DEC) where it simultaneously learns feature representations and cluster assignments using deep neural networks. DEC learns a mapping from the data space to a lower-dimensional feature space in which it iteratively optimizes a clustering objective. In this process, everytime, a new set of images belonging to a class arrives, the autoencoder should be trained again including the new set of arrived images.

A new technique deep transfer clustering \cite{han2019learning} uses prior knowledge of related but different classes to reduce the ambiguity of clustering. The algorithm is improvised by introducing a representation bottleneck, temporal ensembling, and consistency. This also transfers knowledge from the known classes, using them as probes to diagnose different choices for the number of classes in the unlabelled subset. The retraining process to extract the features from a set of images belonging to a new class becomes redundant. But, this methodology results can be improvised for the image collection where the distribution is less variant.

Visual Explanations is a technique where we can visually validate what the model is looking at, verifying that it is indeed looking at the correct patterns in the image and activating around those patterns. If the model is not activating around the proper patterns/objects in the image, then we can revisit the process of training for required feature extraction. Grad-CAM is a tool \cite{8237336}, which is used to visualize the flow of gradients through a neural network so that we get to know what the model is actually learning. When the data distribution is less variant, the features to be extracted for clustering should be differentiable which can enhance the performance of clustering.

With this regard of the model to be able to learn those differentiable features and to concentrate on the required/instructed region of interest in the images, we propose this methodology to train the model for feature extraction which serves the purpose. Our method is an add-on to the existing deep transfer clustering method \cite{han2019learning} which helps the feature extractor to learn more robust and differentiable features rather than concentrating on background common features.

When we use GradCAM, GradCAM++ \cite{8237336}\cite{Chattopadhay_2018} for visualizing what the model is learning in case of classifiers, it can be observed that the model is trying to learn only a part of the image in the dataset. But when such classifiers are used as the initial feature extractors for the deep transfer clustering where the inter-class variance is very less, the clustering performance decreases. So to overcome this, we introduce manual attention mechanism while training the classifiers for the labelled data which not only makes the feature vectors more robust to the region of interest, but also differentiable features are paid more attention. We introduce the manual attention in the training process of classifier by modifying the images, i.e., the unwanted background part of the images are discarded. GrabCut \cite{rother2004grabcut}, is one such algorithm which is used for the removal of the unwanted background part of the image.

\begin{figure}[h]
\begin{center}

\includegraphics[width=10cm,height=5cm]{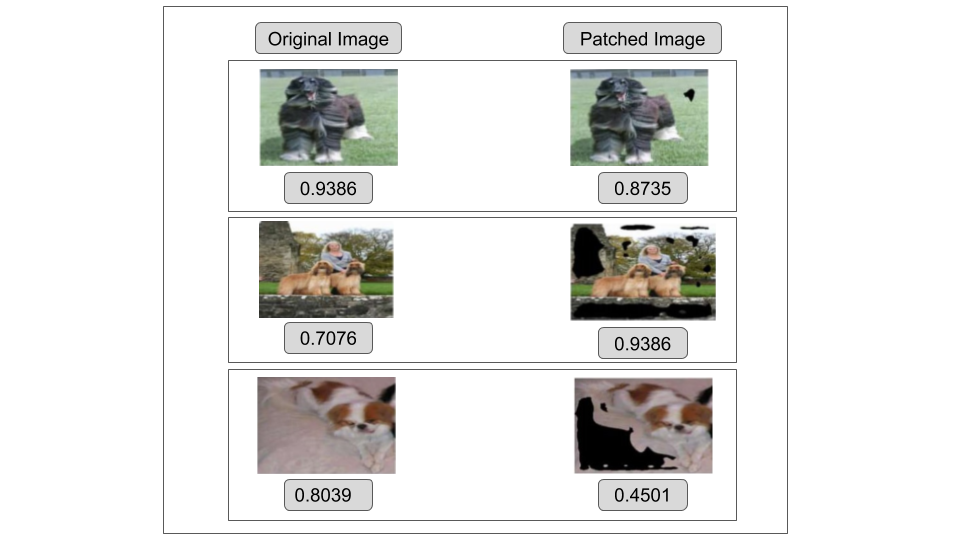}
\caption{Uncertain behaviour in the prediction accuracy of the classifier by adding a patch in certain background region of the images.}
\label{fig:one}
\end{center}
\end{figure}

Features cannot be extracted using regular classifiers
which are trained on the raw images, in clustering process
where the data is less variant, because providing the raw
images to the classifier does not imply what the model
should focus to learn. This results in uncertain behaviour of the model when a
small region is patched over the image as can be observed
in the Figure \ref{fig:one}.

\begin{figure}[h]
\begin{center}
\includegraphics[scale=0.40]{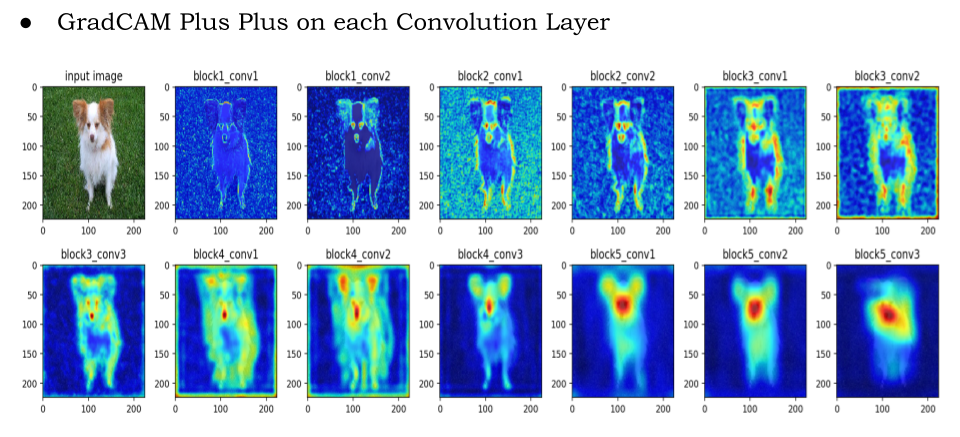}
\caption{Visualization of the each convolution layer of VGG16 classifier using GradCAM++.}
\label{fig:two}
\end{center}
\end{figure}

By visualizing the model using some visualizing algorithms
such as GradCAM, GradCAM++, it can be observed that
the model concentrates on the part of the object rather on
the important aspects of the image which is important in
case of intra-class clustering as depicted in Figure. \ref{fig:two}.


\newpage
\section{Related Work}
\subsection{Deep Adaptive Clustering}\label{A}
Deep adaptive clustering (DAC) \cite{chang2017deep} recasts the clustering problem into a binary pairwise-classification framework to judge whether pairs of images belong to the same clusters. In DAC, the similarities are calculated as the cosine distance between label features of images which are generated by a deep convolutional network (ConvNet). By introducing a constraint into DAC, the learned label features tend to be one-hot vectors that can be utilized for clustering images. The main challenge is that the ground-truth similarities are unknown in image clustering. This issue is handled by presenting an alternating iterative adaptive learning algorithm where each iteration alternately selects labeled samples and trains the ConvNet.

\subsection{Deep Embedded Clustering}\label{AAA}
Deep embedded clustering (DEC) \cite{DBLP:journals/corr/XieGF15} is the algorithm that clusters a set of data points in a jointly optimized feature space. DEC works by iteratively optimizing a KL divergence-based clustering objective with a self training target distribution. This method can be viewed as an unsupervised extension of semi-supervised self-training. This framework provides a way to learn a representation specialized for clustering without ground truth cluster membership labels. DEC offers improved performance as well as robustness for hyperparameter settings, which is particularly important in unsupervised tasks since cross-validation is not possible. 

\subsection{Deep Transfer Clustering}\label{AA}
Deep transfer clustering \cite{han2019learning} which is the improvisation of the DEC has introduced a simple and effective approach for novel visual category discovery in unlabelled data. This method can simultaneously learn data representation and cluster the unlabelled data of novel visual categories while leveraging the knowledge of related categories in labeled data. This method also proposes a novel method to reliably estimate the number of categories in unlabelled data by transferring cluster prior knowledge using labeled probe data. This method overcomes the additional training of the autoencoder with old and new classes. This method uses a classifier as a feature extractor instead of the autoencoders. Loss functions of deep embedded clustering are imposed with certain constraints and temporal ensembling methods to get better results.

\vspace{5pt}

All the state-of-the-art methods shows the results on the datasets (MNIST, Omniglot, CIFAR10, CIFAR100) where the class variance is very high. In this paper we show the results of our methodology on the dog dataset (a subset of Imagenet dataset)
where the data distribution is less variant, experiments show that our improvements perform better than deep transfer clustering (DTC).

\section{Deep Visual Attention Based Transfer Clustering}

    \begin{figure}[h]
    \begin{center}
    \includegraphics[scale=0.10]{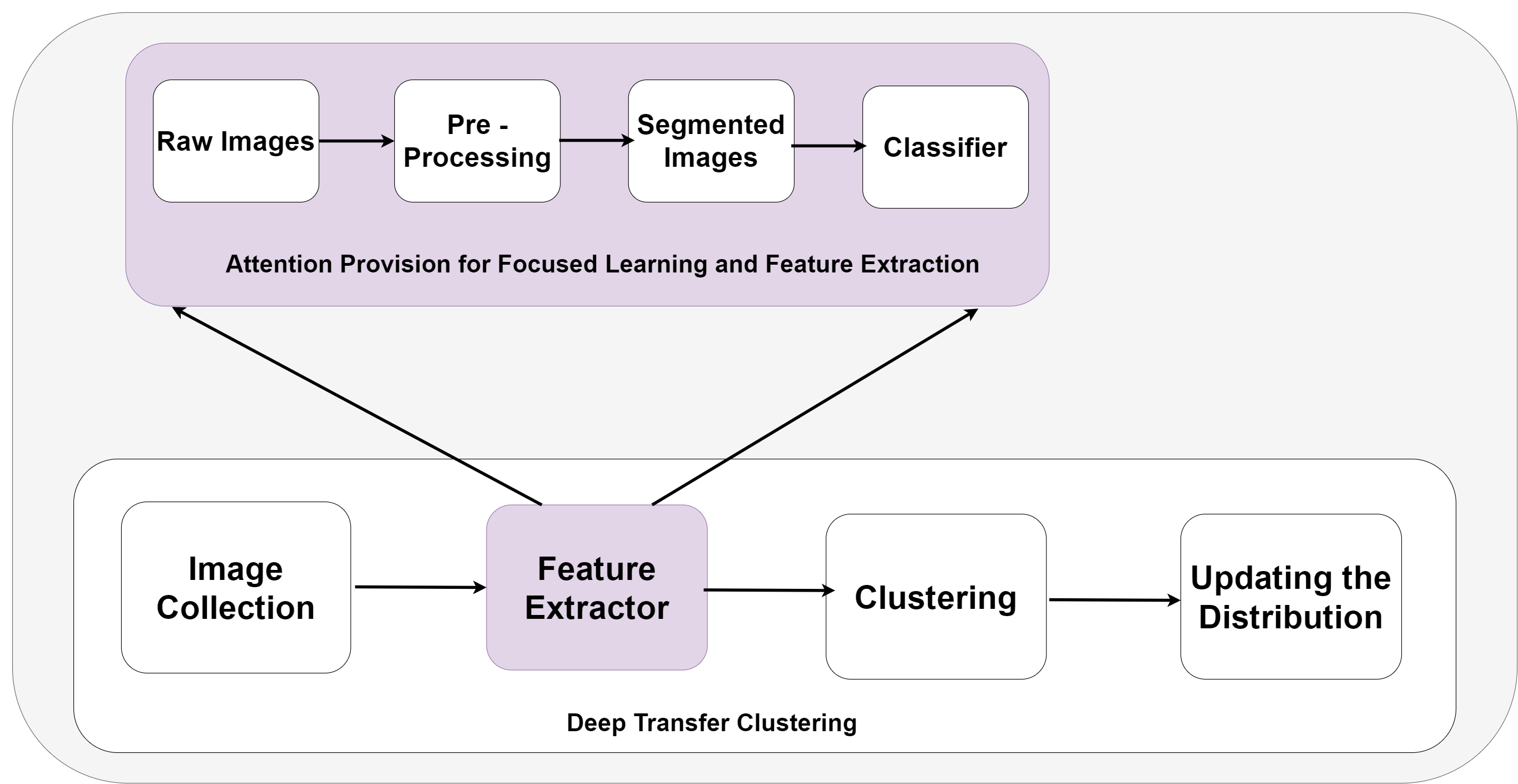}
    \caption{Deep visual attention based transfer clustering}
    \label{fig:three}
    \end{center}
    \end{figure}

Our methodology is the improvement of deep transfer clustering (DTC) for intra-class clustering. A major step in DTC, i.e., training a classifier that is used as a feature extractor is improvised to get more robust and differentiable features. As we can see in the Fig.2 that the classifiers learn some background instead of focusing more on important aspects of the image. This might be good for the classification where training and testing classes are the same, but in the case of the clustering of the intra-classes, this might affect the performance of clustering. So instead of providing the raw images for the training of the classifier, images where the background part is removed and only the part of the image which can help the model to focus on the differentiable features is provided as shown in Fig.3.

\vspace{6pt}
    So, to remove the background from the image, a tool using the GrabCut Algorithm, which can instruct the model to focus on the region of interest for data preprocessing is designed as shown in the Fig.4. This tool helps to get the ROI (region of interest) in an image. GrabCut is an image segmentation method based on graph cuts. It takes a user-specified bounding box around the object to be segmented, and the algorithm estimates the color distribution of the target object and that of the background using a Gaussian mixture model.

    \begin{figure}
    \begin{center}
    \includegraphics[scale=0.25]{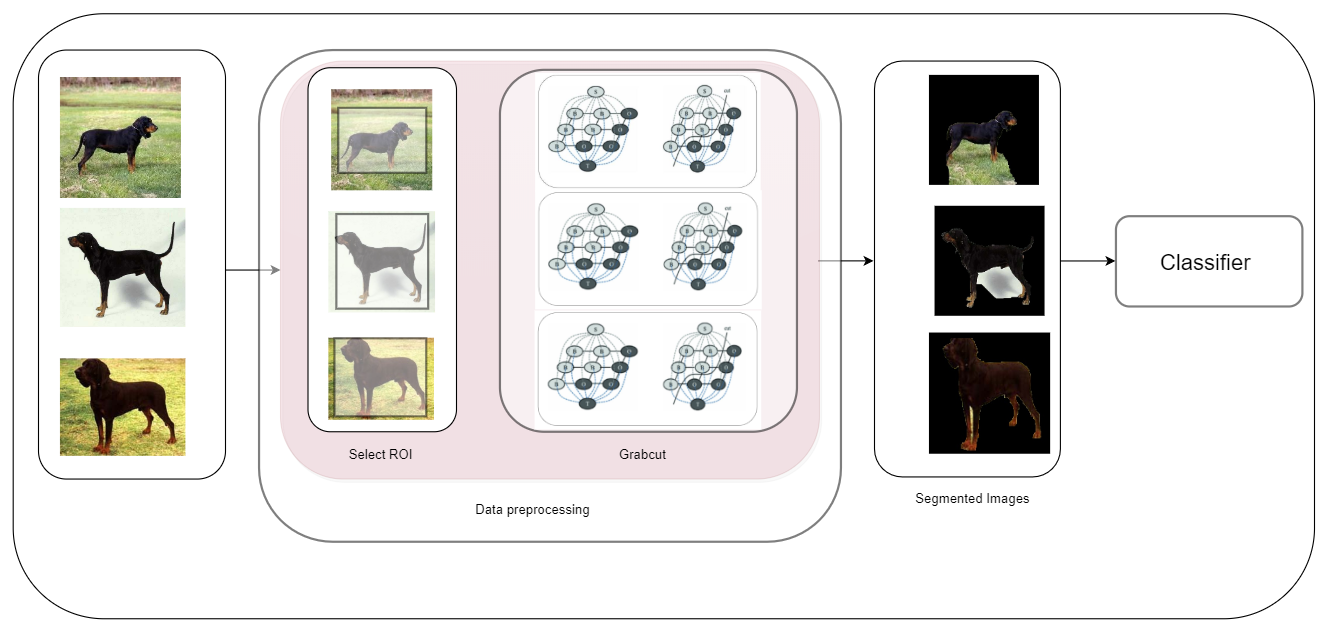}
    \caption{Our method of pre-processing the data before training the classifier for the initial feature extraction}
    \label{fig:four}
    \end{center}
    \end{figure}

\begin{itemize}
    \item {We train the classifier with the background-subtracted images. Background subtraction is done using the tool which we have created using the grabcut algorithm. Now, we have the feature extractor for the upcoming new classes. Though the features may be good for new classes, it might not be the best.}
    \vspace{5pt}
    \item {Extract the features for the new unlabelled images using the previously trained classifier model, reduce the dimensions to the $K$ dimensions using the PCA and initialize the cluster centers $U$ using the k-means and incorporate the PCA as the last linear layer in the classifier model, 
    where 
    $
        U = \{ \mu_k,k = 1....K\}
    $
        represents  the  cluster centers, $f_\theta$  represents the feature extractor.}
    \vspace{5pt}    
    \item {Train the $\theta$ of the feature extractor  $f_\theta$ and cluster centers $U$ using q as the target distributions and update the target distribution $q$ . Here, the loss function is combination of DEC loss and the consistency constraints loss  which is as follows.
    Let $p(k|i)$ be the probability of assigning data point $i \in  \{1...N\}$ to cluster $k \in \{1...K\}$.
    \begin{equation}
    \label{eq:1}
        p(k|i) 	\propto (1+\frac{\parallel z_i + \mu_k \parallel^2}{\alpha})^{-\frac{\alpha+1}{2}}
    \end{equation}
    \item{Further assuming that data indices are sampled uniformly (i.e $p(i)$ = 1/N), we can write the joint distribution as $p(i,k) = p(k|i)/N$. 
    DEC defines its objective as KL divergence loss between the soft assignments $q$ and auxiliary distribution $p$ as follows}
    \begin{equation}
    \label{eq:2}
        E(q) = KL(q\parallel p) = \frac{1}{N} \sum_{i=1}^{N}\sum_{k=1}^{K}q(k|i) \log {\frac{q(k|i)}{p(k|i)}}
    \end{equation}
    \item{We minimize the KL divergence between the auxiliary distribution and target distribution to attain the good solution. we have the formula to calculate the auxiliary distribution $p$. Target distribution $q$ is calculated as follows:}
    \begin{equation}
    \label{eq:3}
        q(k|i) \propto p(k|i) \cdot p(i|k)
    \end{equation}
    \item{In this manner, the assignment of image $i$ to cluster $k$ is reinforced when the current distribution assigns a high probability of going from $i$ to $k$ as well as of going from $k$ to $i$. The latter has an equalization effect as the probability of sampling data point $i$ in cluster $k$ is high only if the cluster is not too large. Using Bayes rule for $p(k|i)$ as follows:}
    \begin{equation}
    \label{eq:4}
        q(k|i) \propto \frac{p(k|i)^2}{\sum_{i=1}^{N}p(k|i)}
    \end{equation}
    
    \item{DTC uses the above loss and also the additional function in the loss i.e., consistency constrain. Consistency constrains have shown the effective results in case of semi-supervised learning.
    A consistency constraint can be incorporated by enforcing the predictions of a data sample and its transformed counterpart (which can be obtained by applying data transformation such as random cropping and horizontal flipping on the original data sample).}
    Hence the final loss function is defined as:
    \begin{equation}
    \label{eq:5}
        E(q) = L_1 + L_2
    \end{equation}
    \begin{equation}
    \label{eq:6}
        L_1 = \frac{1}{N} \sum_{i=1}^{N}\sum_{k=1}^{K}q(k|i) \log {\frac{q(k|i)}{p(k|i)}}
    \end{equation}
    \begin{equation}
    \label{eq:7}
        L_2 = \omega(t)\frac{1}{NK}\sum_{i=1}^{N}\sum_{k=1}{K}\parallel p(k|i) - p'(k|i)\parallel
    \end{equation}
    
    }

    \vspace{6pt}
    where $p'(k|i)$ is either the prediction of the transformed sample or the temporal ensemble prediction
    $p^{\sim t}(k|i)$, and $\omega(t)$ is a ramp-up function as used to gradually increase the weight of the consistency constraint from 0 to 1.
    \vspace{6pt}

    \item{Predict the probability of assignment $p(k|i)$ of image $i \in  \{1...N\}$ to the cluster  $k \in \{1...K\}$ using the equation and get the index of the cluster to each image with the highest $p(k|i)$}
    
    DTC improves DEC smoothing of the cluster assignment via temporal ensembling
    i.e.,
    the clustering model's $p$ is computed at different epochs are aggregated by maintaining an exponential moving average (EMA) of the previous distributions.
    \begin{equation}
        P^t(k|i) = \beta \cdot p^{t-1}(k|i) + (1 - \beta) \cdot p^t(k|i).
    \end{equation}
    where $\beta$ is a momentum term controlling how far the ensemble reaches into training history, and $t$ indicates the time step. To correct the zero initialization of the EMA, $P^t$ is rescaled to obtain the smoothed model distribution i.e.,
    \begin{equation}
    \label{eq:8}
        p^{\sim t}(k|i) = \frac{1}{1 - \beta ^ t} . P^t(k|i)
    \end{equation}
\end{itemize}

The concepts discussed in above Equations. \ref{eq:1},\ref{eq:2},\ref{eq:3},\ref{eq:4},\ref{eq:5},\ref{eq:6},\ref{eq:7},\ref{eq:8} are inspired from \cite{han2019learning}.

    DTC has improved the DEC algorithm by using Temporal ensembling and consistency constraints, we show that using our method of training the initial feature extractor by using the manual attention over the training the images gives better results than the regular classifiers for the data where there is less variance between the classes, we have changed the initial training process of the DTC, the remaining algorithm remains same as that of DTC.
    
\begin{figure}
    \begin{center}
    \includegraphics[width=14cm,height=6cm]{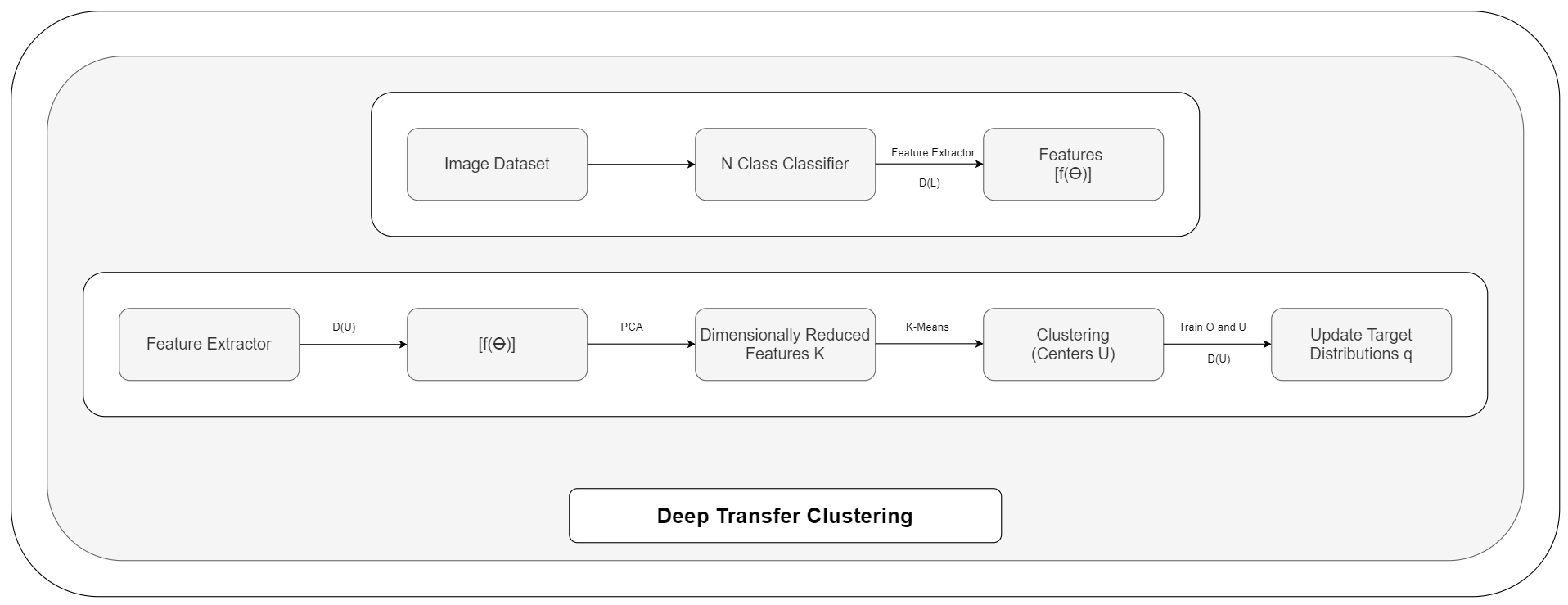}
    \caption{Deep transfer clustering}
    \label{fig:five}
    \end{center}
    \end{figure}    
    
    Figure. \ref{fig:five} is the complete algorithm of the deep transfer clustering (DTC), which includes two major steps training of classifier and the fine tuning of the feature extractor model with respect to the cluster assignmet loss and other constraints, we have changed the first step of the DTC algorithm which is best suitable for the Intra-class clustering.
    
\section{Results}
    The results were generated for the dog dataset, the subset of the imagenet dataset. This dataset was chosen because it has very high similarities between the classes, and also in many of the dog classes dogs face of one class is somewhat similar to other classes, so we need our classifier to concentrate on the whole dog in the image rather than just concentrating on the face of the dog which is always true in case of the classifier trained using raw images. Our improvisation was compared with DTC.
    
    DTC has three different algorithms namely:
    \begin{itemize}
    \item{DTC-PI: Model trained using DEC loss with consistency constraint between predictions of a sample and its transformed counterpart.}

\vspace{10pt}
\hspace{0pt}
\begin{tabular}{ |p{3cm}|p{2cm}|p{5cm}|  }
\hline
\multicolumn{3}{|c|}{Pi model} \\
\hline
 & DTC &Our proposed model \\
\hline
Dataset & Dog &Dog \\
Accuracy &  0.7784  & 0.7791 \\
NMI &0.7369 &0.7378\\
ARI    &0.6458 &0.6466\\
\hline
\end{tabular}
\vspace{5pt}

The above results were generated for the DTC-PI model, where loss function has the DEC loss and the consistency constraint, our results are slightly better than the original DTC model's accuracy.

\vspace{9pt}
    \item{DTC-Baseline: Model trained using the just DEC loss.}

\hspace{0pt}
\begin{tabular}{ |p{3cm}|p{2cm}|p{5cm}|  }
\hline
\multicolumn{3}{|c|}{Baseline model} \\
\hline
 & DTC &Our proposed model \\
\hline
Dataset & Dog &Dog \\
Accuracy &0.7508 &0.7811\\
NMI &0.7172 &0.7481\\
ARI   &0.6442 &0.6668\\
\hline
\end{tabular}
\vspace{5pt}

The above results were generated for the DTC-Baseline model, where the loss function is just the DEC loss, our results are 3\% higher in case of accuracy and NMI and 4\% higher in the case of ARI.

\vspace{5pt}
\item{DTC-TEP: Model trained using DEC loss with consistency constraint between the current prediction and temporal ensemble prediction of each sample.}
    \vspace{5pt}

\vspace{5pt}

\hspace{0pt}
\begin{tabular}{ |p{3cm}|p{2cm}|p{5cm}|  }
\hline
\multicolumn{3}{|c|}{TEP model} \\
\hline
 & DTC &Our proposed model \\
\hline
Dataset & Dog &Dog \\
Accuracy &0.7534 &0.7779\\
NMI &0.7211 &0.7455\\
ARI    &0.6285 &0.6615\\
\hline
\end{tabular}
\vspace{5pt}

The above results were generated for the DTC-TEP model, where loss function consists of DEC loss and consistency constraint, the auxiliary distribution $p(k|i)$ is calculated using the temporal ensembling method. Our results are 2.5\% higher in the case of accuracy and NMI 4\% higher in the case of ARI.
\end{itemize}

\newpage
\section{Conclusion}
In this paper, various facts could be concluded, some of them including that the traditional clustering techniques of using pre-trained models for image clustering will not lead to better results for all the kinds of data, especially for those classes where the similarity between the two classes is very high. In such cases where the dataset contains the classes in which it is very difficult to differentiate between the classes. It is very difficult to extract the differentiable features from the dataset.

A simple and effective approach for the categorization of unlabelled data is introduced, by considering it as a deep transfer clustering problem. The method proposed can simultaneously learn to extract the features that can differentiate between two classes, learn a data representation and cluster the unlabelled data of novel visual categories, while leveraging the knowledge of related categories in labeled data. The main focus was to cluster the image collection where the similarity between the two classes is very high. This we have achieved by plugging in a module that makes the network concentrate on the instructed region of interest rather than a complete image.

\bibliographystyle{splncs04}
\bibliography{Biblography}

\end{document}